\documentclass[letterpaper]{article}
\usepackage[preprint]{aaai2027}

\usepackage[hyphens]{url}
\usepackage{graphicx}
\usepackage{natbib}
\usepackage{caption}
\usepackage{amsmath}
\usepackage{amssymb}
\usepackage{algorithm}
\usepackage{algorithmic}
\usepackage{booktabs}
\usepackage{multirow}
\usepackage{amsthm}

\begin{document}
%
\title{Low-Dimensional High-Leverage Subspace Optimization: Beyond Full-Parameter Coupled Training for Neural Network Quantization}
\author{
Peng Xia\textsuperscript{1},
Junbiao Pang\textsuperscript{1}\corresponding,
Zheng Huang\textsuperscript{1}
}
\affiliations{
\textsuperscript{1}School of Information Science and Technology, Beijing University of Technology, Beijing, China\\
\{xiapeng@emails.bjut.edu.cn, junbiao\_pang@bjut.edu.cn\}
}

\maketitle
\begin{abstract}
Low-bit quantization suffers severe accuracy degradation on compact networks, rooted in the dominant full-parameter coupled training paradigm that ignores parameter subspace heterogeneity. Their limited feature redundancy leaves little room to absorb quantization errors. Conventional pipelines adopt monolithic optimization: PTQ reconstructs fixed pretrained models without improving inherent quantization friendliness; QAT updates all parameters jointly, suffering from gradient coupling between backbone weights and calibration parameters. In this paper, we identify normalization affine parameters as a low-dimensional high-leverage subspace dominating quantization robustness, and propose Normalization Affine Preconditioning (NAP) for targeted subspace optimization. For PTQ, NAP freezes backbone weights and fine-tunes only affine parameters under the target fake-quantization graph on full-precision models, proactively boosting quantization friendliness before downstream reconstruction. For QAT, we introduce an alternating QAT-NAP schema that decouples feature learning and numerical calibration, breaking the performance ceiling of saturated joint training. Theoretical analysis confirms BN affine parameters fully cancel the channel-wise affine component of quantization distortion, while nonlinear rounding and clipping residuals form the irreducible error boundary; distillation-guided NAP acts as directional flatness optimization, projecting teacher-student logit mismatch onto the restricted subspace. Experiments on ImageNet and CIFAR-100 show NAP recovers severely collapsed low-bit quantization, consistently boosts reconstruction-based PTQ, and outperforms saturated full-parameter QAT with negligible tuning cost. This work reveals the principle of targeted low-dimensional subspace optimization, offering a new perspective beyond full-parameter coupled training for efficient deep learning.

\end{abstract}

\section{Introduction}

Post-training quantization (PTQ) compresses neural networks by converting
full-precision weights and activations into low-bit representations without
costly end-to-end retraining. Conventional PTQ pipelines calibrate
quantization ranges or equalize layer responses before quantization
~\cite{jacob2018quantization,nagel2019dfq}. Recent reconstruction-based
methods further improve accuracy by learning rounding decisions
~\cite{nagel2020adaround}, reconstructing blocks~\cite{li2021brecq},
randomly dropping activation quantization during optimization
~\cite{wei2022qdrop}, or compensating prediction differences
~\cite{liu2023pdquant}. These techniques have substantially improved
moderate-bit quantization, but extreme low-bit weight--activation quantization
remains brittle. The problem is particularly severe for compact models such
as MobileNetV2~\cite{sandler2018mobilenetv2}, whose depthwise-separable
structure and limited channel redundancy amplify clipping, rounding, and
scale mismatch~\cite{nagel2022oscillations,wei2022qdrop}.

\begin{figure}[t]
\centering
\includegraphics[width=1\linewidth]{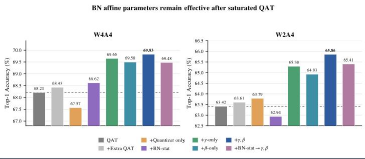}
\caption{
\textbf{Diagnostic study after saturated LSQ+ QAT.}
All adaptation branches start from the same LSQ+ QAT checkpoint on
CIFAR-100 MobileNetV2. The FP32 accuracy of the pretrained model is
71.3\%.
}
\label{fig:qat_diagnostic}
\end{figure}

Most quantization pipelines implicitly treat all trainable parameters as
equally useful adaptation variables. Our diagnostic in
Fig.~\ref{fig:qat_diagnostic} challenges this assumption. Starting from the
same saturated LSQ+ checkpoint, matched extra QAT provides only marginal
improvement, while quantizer-only tuning can even degrade performance. In
contrast, updating only BN affine parameters improves W4A4 accuracy from
68.21\% to 69.83\% and W2A4 accuracy from 63.42\% to 65.86\%. This branch
updates only 34.2K parameters, approximately 1.43\% of the full-QAT trainable
state. The result indicates that normalization affine parameters constitute
an under-exploited, low-dimensional response-control subspace rather than an
arbitrary sparse parameter subset.

Motivated by this observation, we propose \textbf{N}ormalization
\textbf{A}ffine \textbf{P}re-conditioning (NAP). NAP freezes convolutional
and linear weights and adapts only the scale and shift parameters already
present in normalization layers. For PTQ, the adaptation is performed under
the same fake-quantized graph used for evaluation or downstream
reconstruction. For saturated QAT checkpoints, NAP provides a lightweight
post-QAT adaptation stage. For RMSNorm-based language models, we further
alternate between normalization-scale updates and group-wise
quantization-scale updates while keeping the backbone frozen. This formulation
separates feature parameters, response-control parameters, and discretization
parameters instead of optimizing them monolithically.

Our analysis explains both the leverage and the limits of this subspace. BN
affine parameters act as channel-wise gain and offset controllers, while
RMSNorm retains only the gain term~\cite{ioffe2015batchnorm,zhang2019rmsnorm}.
Under a local affine-distortion model, these parameters can compensate the
structured scale-and-shift component of quantization error, whereas nonlinear
rounding, clipping, and sample-dependent residuals remain outside the affine
subspace. From the prediction perspective, distillation-guided NAP solves a
local projection problem: the teacher--student mismatch defines the desired
correction and the normalization-affine Jacobian determines which component
is expressible without changing backbone weights.

Our contributions are summarized as follows:
\begin{itemize}
\item We systematically characterize normalization affine parameters as a
low-dimensional, high-leverage response-control subspace for low-bit
quantization across BN- and RMSNorm-based architectures.

\item We propose a target-aligned NAP framework supporting pre-conditioning
before PTQ, affine adaptation after saturated QAT, and alternating
optimization of normalization responses and quantization scales.

\item We provide a residual-aware local analysis that distinguishes
affine-compensable distortion from irreducible rounding and clipping
residuals, and interprets distillation-guided NAP as projection onto the
normalization-affine response subspace.

\item Experiments on ImageNet, CIFAR-100, Cityscapes, and
Qwen2.5-3B-Instruct evaluate NAP across classification, semantic
segmentation, and language modeling, while controlled studies identify its
backend sensitivity, data dependence, and failure cases.
\end{itemize}

\section{Related Work}

\subsection{Post-Training Quantization and Quantization-Aware Training}

PTQ reduces model precision without end-to-end retraining. Early methods
calibrate clipping ranges, correct bias, or equalize channel responses
~\cite{jacob2018quantization,nagel2019dfq}. Reconstruction-based methods
subsequently improve low-bit accuracy by learning rounding decisions or
matching intermediate representations. AdaRound~\cite{nagel2020adaround}
optimizes weight rounding, BRECQ~\cite{li2021brecq} performs block-wise
reconstruction, QDrop~\cite{wei2022qdrop} stochastically disables activation
quantization during reconstruction, and PD-Quant~\cite{liu2023pdquant}
introduces prediction-aware objectives.

QAT instead exposes the model to fake quantization during training.
Representative methods learn clipping ranges, smooth the discrete quantizer,
or optimize step sizes and offsets, including PACT~\cite{choi2018pact},
DSQ~\cite{gong2019dsq}, LSQ~\cite{esser2020lsq}, and
LSQ+~\cite{bhalgat2020lsqplus}. Weight oscillations can destabilize low-bit
QAT, motivating oscillation dampening and iterative weight freezing
~\cite{nagel2022oscillations}; StableQAT develops a bounded Fourier-based
surrogate for ultra-low-bit optimization~\cite{chen2026stableqat}. NAP is
complementary to these approaches: it isolates the normalization affine
subspace before PTQ, after saturated QAT, or between quantization-scale
updates.

\subsection{Quantization-Friendly Transformations}

A complementary line of work changes the parameterization observed by the
quantizer. Cross-layer equalization redistributes channel ranges without
changing the represented function~\cite{nagel2019dfq}. SmoothQuant transfers
activation-outlier difficulty to weights through equivalent channel-wise
scaling~\cite{xiao2023smoothquant}, while OmniQuant learns clipping
thresholds and equivalent transformations under block-wise reconstruction
~\cite{shao2024omniquant}. Rotation-based methods such as QuaRot and
SpinQuant suppress outliers using invariant random or learned rotations
~\cite{ashkboos2024quarot,liu2025spinquant}. FlatQuant further learns
layer-specific affine transformations for full weight--activation
quantization~\cite{sun2025flatquant}. These methods introduce auxiliary
scales, rotations, or transformation matrices. NAP instead optimizes affine
parameters already present in BN or RMSNorm and therefore adds no inference
branch.

\subsection{Large Language Model Quantization}

LLM quantization has developed along weight-only, weight--activation, and QAT
directions. GPTQ uses approximate second-order information for one-shot
weight-only quantization~\cite{frantar2023gptq}, whereas AWQ protects
activation-salient weight channels through equivalent scaling
~\cite{lin2024awq}. SmoothQuant targets hardware-friendly W8A8 inference
~\cite{xiao2023smoothquant}, while OmniQuant, QuaRot, SpinQuant, and FlatQuant
extend accurate quantization toward lower-bit weight--activation settings
~\cite{shao2024omniquant,ashkboos2024quarot,liu2025spinquant,sun2025flatquant}. LLM-QAT uses data-free distillation to train low-bit
weights, activations, and KV caches~\cite{liu2023llmqat}; EfficientQAT lowers
training cost through block-wise all-parameter optimization followed by
end-to-end quantization-parameter tuning~\cite{chen2025efficientqat}.

The closest prior work to our RMSNorm experiments is Norm Tweaking
~\cite{li2024normtweaking}, which updates normalization-layer weights to
align quantized and full-precision activation distributions for LLM PTQ.
Accordingly, NAP does not claim to be the first normalization-only
quantization method. Its distinction is a unified normalization-affine
formulation across BN-based CNNs and RMSNorm-based LLMs, explicit alignment
with the target fake-quantized graph, use before PTQ and after saturated QAT,
and alternating response/quantization-grid optimization. The residual-aware
analysis and matched parameter-subspace controls further characterize when
this intrinsic affine subspace is effective and where it fails.

\subsection{Low-Dimensional Adaptation Subspaces}

Parameter-efficient adaptation suggests that small parameter subsets can
exert disproportionately large control over network behavior. Normalization
affine parameters are especially relevant because each gain or bias
coefficient is broadcast over an entire channel. BN provides learnable scale
and shift parameters~\cite{ioffe2015batchnorm}, whereas RMSNorm retains only a
learnable scale~\cite{zhang2019rmsnorm}. Prior normalization-only studies
primarily target domain adaptation, fine-tuning efficiency, or LLM PTQ. NAP
formulates these parameters as a unified response-control subspace for CNN and
LLM quantization: unlike full QAT, it freezes backbone weights; unlike
quantizer-only tuning, it changes the responses presented to the quantizer;
and unlike static reparameterization, it is optimized under the target
backend graph.

\section{Method}
\label{sec:method}

\subsection{Overview and Problem Formulation}
\label{sec:overview}

Let $f(\boldsymbol{x};\mathbf{W},\boldsymbol{\Phi})$ denote a pretrained
network, where $\mathbf{W}$ contains convolutional and linear weights and
$\boldsymbol{\Phi}$ contains the affine parameters of normalization layers.
For BN~\cite{ioffe2015batchnorm}, $\boldsymbol{\Phi}$ contains the
channel-wise scale and shift parameters
$(\boldsymbol{\gamma},\boldsymbol{\beta})$; for
RMSNorm~\cite{zhang2019rmsnorm}, $\boldsymbol{\Phi}$ contains only the scale
parameters $\boldsymbol{\gamma}$. We use
\begin{equation}
\boldsymbol{\Phi}
=
\left\{
\boldsymbol{\gamma}_{l},
\boldsymbol{\beta}_{l}
\right\}_{l=1}^{L},
\label{eq:affine_set}
\end{equation}
with $\boldsymbol{\beta}_{l}$ omitted for normalization layers without a
learnable shift.

NAP is based on a simple decomposition of model adaptation. Backbone weights
$\mathbf{W}$ determine the learned features, quantization parameters
$\boldsymbol{\Omega}$ determine the discretization grid, and normalization
affine parameters $\boldsymbol{\Phi}$ regulate the channel-wise responses seen
by the quantizers. Instead of updating all three parameter groups jointly, NAP
isolates the low-dimensional response-control subspace
$\boldsymbol{\Phi}$. This design gives rise to three uses considered in this
paper: (i) pre-conditioning before PTQ, (ii) affine adaptation after a
saturated QAT checkpoint, and (iii) alternating optimization of
normalization affine parameters and quantization scales.

\begin{figure}[t]
\centering
\includegraphics[width=0.95\linewidth]{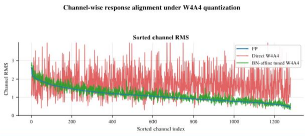}
\caption{
\textbf{Target-aligned normalization affine pre-conditioning.}
The full-precision teacher is fixed, while the fake-quantized student updates
only normalization affine parameters. The adapted channel responses reduce
the scale and offset mismatch presented to the target quantization backend.
}
\label{fig:feature_alignment}
\end{figure}

\subsection{Target-Aligned NAP for PTQ}
\label{sec:nap_ptq}

Let $\mathcal{G}_{\mathcal{B}}$ denote the fake-quantized graph induced by a
target backend $\mathcal{B}$, including quantization insertion points,
observers, granularity, clipping rules, and weight/activation quantizers. A
uniform fake quantizer is written as
\begin{equation}
\mathcal{Q}(\boldsymbol{u};s,z)
=
s\left[
\operatorname{clip}
\left(
\left\lfloor \frac{\boldsymbol{u}}{s}\right\rceil+z,
q_{\min},q_{\max}
\right)-z
\right],
\label{eq:uniform_quantizer}
\end{equation}
where $s$ and $z$ are the scale and zero-point.

After calibration on $\mathcal{D}_{c}$, the quantization parameters are fixed
as $\boldsymbol{\Omega}_{0}$. The target fake-quantized student and the fixed
full-precision teacher are
\begin{align}
\boldsymbol{z}_{q}
&=
\widetilde{f}_{q}
\left(
\boldsymbol{x};
\mathbf{W}_{0},
\boldsymbol{\Phi},
\boldsymbol{\Omega}_{0},
\mathcal{G}_{\mathcal{B}}
\right),
\label{eq:student}\\
\boldsymbol{z}_{0}
&=
f
\left(
\boldsymbol{x};
\mathbf{W}_{0},
\boldsymbol{\Phi}_{0}
\right).
\label{eq:teacher}
\end{align}
NAP freezes convolutional and linear weights and updates only the normalization
affine parameters:
\begin{equation}
\nabla_{\mathbf{W}}\mathcal{L}_{\mathrm{NAP}}=\mathbf{0},
\qquad
\nabla_{\boldsymbol{\Phi}}\mathcal{L}_{\mathrm{NAP}}\neq\mathbf{0}.
\label{eq:freeze_backbone}
\end{equation}

We use a task loss together with prediction-level knowledge
distillation~\cite{hinton2015distilling}. The task loss
is classification cross-entropy for image classification, pixel-wise
cross-entropy for semantic segmentation, and next-token negative
log-likelihood for language modeling. Let
$p_T(\boldsymbol{z})=\operatorname{softmax}(\boldsymbol{z}/T)$. The NAP
objective is
\begin{equation}
\mathcal{L}_{\mathrm{NAP}}
=
\mathcal{L}_{\mathrm{task}}(\boldsymbol{z}_{q},y)
+
\lambda_{\mathrm{KD}}T^2
\operatorname{KL}
\left(
p_T(\boldsymbol{z}_{0})
\Vert
p_T(\boldsymbol{z}_{q})
\right).
\label{eq:nap_objective}
\end{equation}
The task term keeps the fake-quantized model predictive, while the
distillation term preserves the decision structure of the fixed teacher. In
our main low-bit experiments, we additionally use a weak channel-scale
regularizer
\begin{equation}
\mathcal{L}_{\mathrm{CS}}
=
\frac{1}{L}
\sum_{l=1}^{L}
\operatorname{Var}_{c}
\left[
\log\left(|\gamma_{l,c}|+\epsilon\right)
\right],
\label{eq:channel_scale_reg}
\end{equation}
which discourages a few normalization channels from dominating the activation
range. The implemented objective is therefore
$\mathcal{L}_{\mathrm{NAP}}+\lambda_{\mathrm{CS}}\mathcal{L}_{\mathrm{CS}}$,
with a small coefficient specified in the experimental protocol. Crucially, all
terms are optimized under the same graph $\mathcal{G}_{\mathcal{B}}$ used for
evaluation or downstream reconstruction. Changing the graph changes both the
quantization residual and the directions through which $\boldsymbol{\Phi}$
can correct it; therefore a NAP checkpoint is not expected to transfer
perfectly across mismatched backends.

\subsection{NAP after QAT and Alternating NAP-QAT}
\label{sec:nap_qat}

NAP can also be applied to a saturated QAT checkpoint
$(\mathbf{W}_{\mathrm{qat}},\boldsymbol{\Phi}_{\mathrm{qat}},
\boldsymbol{\Omega}_{\mathrm{qat}})$. In this setting, the backbone and
quantization parameters are fixed and only the normalization affine
parameters are adapted:
\begin{equation}
\boldsymbol{\Phi}^{*}
=
\arg\min_{\boldsymbol{\Phi}}
\mathcal{L}_{\mathrm{NAP}}
\left(
\mathbf{W}_{\mathrm{qat}},
\boldsymbol{\Phi},
\boldsymbol{\Omega}_{\mathrm{qat}}
\right).
\label{eq:post_qat_nap}
\end{equation}
This post-QAT form tests whether a saturated jointly trained model still
contains an under-exploited affine adaptation direction.

For large language models, we additionally use an alternating variant that
decouples quantization-grid adaptation from response adaptation. Backbone
weights remain frozen. We denote by $\mathcal{L}_{q}$ the same
target-graph task-and-distillation objective when it is optimized with
respect to the quantization parameters. At alternating round $r$, we first
update group-wise quantization scales while fixing normalization parameters,
\begin{equation}
\boldsymbol{\Omega}^{r+1}
\leftarrow
\arg\min_{\boldsymbol{\Omega}}
\mathcal{L}_{q}
\left(
\mathbf{W}_{0},
\boldsymbol{\Phi}^{r},
\boldsymbol{\Omega}
\right),
\label{eq:qscale_step}
\end{equation}
and then update RMSNorm scales while fixing the quantization grid,
\begin{equation}
\boldsymbol{\Phi}^{r+1}
\leftarrow
\arg\min_{\boldsymbol{\Phi}}
\mathcal{L}_{\mathrm{NAP}}
\left(
\mathbf{W}_{0},
\boldsymbol{\Phi},
\boldsymbol{\Omega}^{r+1}
\right).
\label{eq:nap_step}
\end{equation}
We refer to this block-coordinate procedure as NAP-QAT. QScale-QAT adapts only
$\boldsymbol{\Omega}$, RMSNorm-NAP adapts only $\boldsymbol{\Phi}$, and
NAP-QAT alternates between the two complementary subspaces. This formulation
also clarifies that the large-model experiments do not perform
full-parameter QAT.

\subsection{Why Is the Affine Subspace High-Leverage?}
\label{sec:mechanism}

\paragraph{Channel-wise broadcast control.}
For a normalized feature channel $\boldsymbol{h}_{l,c}$, the affine response is
\begin{equation}
\boldsymbol{a}_{l,c}
=
\gamma_{l,c}\boldsymbol{h}_{l,c}
+
\beta_{l,c},
\label{eq:affine_response}
\end{equation}
where $\beta_{l,c}=0$ for RMSNorm. A small affine update gives
\begin{equation}
\Delta\boldsymbol{a}_{l,c}
=
\boldsymbol{h}_{l,c}\Delta\gamma_{l,c}
+
\mathbf{1}\Delta\beta_{l,c}.
\label{eq:broadcast_update}
\end{equation}
Thus, one or two scalar parameters modify an entire channel over all spatial
positions or tokens. The parameter space is small, but its response is
broadcast over a high-dimensional activation tensor.

\paragraph{Residual-aware affine compensation.}
Consider a pre-BN activation whose quantized counterpart admits the local
decomposition
\begin{equation}
\widehat{a}_{c}
=
\alpha_{c}a_{c}
+
\delta_{c}
+
e_{c},
\qquad \alpha_{c}>0,
\label{eq:error_decomposition}
\end{equation}
where $\alpha_{c}a_{c}+\delta_{c}$ is the structured channel-wise affine
component and $e_c$ contains rounding, clipping, saturation, and
sample-dependent residuals. Let
$\rho_c=\sqrt{\sigma_c^2+\epsilon}$ and let
$(\gamma_c^0,\beta_c^0)$ be the original BN affine parameters. Choosing
\begin{align}
\gamma_c^{*}
&=
\frac{\gamma_c^0}{\alpha_c},\\
\beta_c^{*}
&=
\beta_c^0
-
\frac{\gamma_c^0}{\alpha_c\rho_c}
\left[
(\alpha_c-1)\mu_c+\delta_c
\right]
\label{eq:compensation}
\end{align}
yields
\begin{equation}
\widehat{y}_c^{*}
=
y_c^0
+
\frac{\gamma_c^0}{\alpha_c\rho_c}e_c.
\label{eq:compensation_residual}
\end{equation}
Therefore BN affine parameters can exactly cancel the scale-and-shift component
under this local model, while nonlinear residuals remain. RMSNorm has no shift
parameter, so it can compensate the multiplicative component but not a general
additive offset. This distinction explains why normalization affine parameters
are high-leverage without claiming that they can recover information destroyed
by severe clipping.

\paragraph{Local projection view.}
Let
$\boldsymbol{e}_{z}
=
\boldsymbol{z}_{0}
-
\boldsymbol{z}_{q}(\boldsymbol{\Phi}_{0})$
be the teacher--student logit mismatch before adaptation. For a small affine
update,
\begin{equation}
\boldsymbol{z}_{q}
(\boldsymbol{\Phi}_{0}+\Delta\boldsymbol{\Phi})
\approx
\boldsymbol{z}_{q}(\boldsymbol{\Phi}_{0})
+
\mathbf{J}_{q}\Delta\boldsymbol{\Phi},
\label{eq:logit_linearization}
\end{equation}
where $\mathbf{J}_{q}$ is the logit Jacobian under the target quantized graph.
Locally, the distillation term seeks
\begin{equation}
\min_{\Delta\boldsymbol{\Phi}}
\left\|
\boldsymbol{e}_{z}
-
\mathbf{J}_{q}\Delta\boldsymbol{\Phi}
\right\|_{\mathbf{H}_{z}}^{2}.
\label{eq:projection}
\end{equation}
NAP therefore corrects the component of the quantization-induced mismatch that
is expressible in the normalization affine response subspace. We treat this as
a local mechanism rather than a claim of globally flatter minima.

\subsection{Algorithm and Complexity}
\label{sec:nap_algorithm}

\begin{algorithm}[t]
\caption{Target-Aligned Normalization Affine Pre-conditioning}
\label{alg:nap}
\begin{algorithmic}[1]
\REQUIRE Pretrained model
$f(\cdot;\mathbf{W}_{0},\boldsymbol{\Phi}_{0})$,
target graph $\mathcal{G}_{\mathcal{B}}$,
calibration set $\mathcal{D}_{c}$,
tuning set $\mathcal{D}_{t}$
\ENSURE Adapted normalization parameters $\boldsymbol{\Phi}^{*}$
\STATE Calibrate $\mathcal{G}_{\mathcal{B}}$ on $\mathcal{D}_{c}$ and fix
$\boldsymbol{\Omega}_{0}$
\STATE Freeze $\mathbf{W}_{0}$ and enable gradients only for
$\boldsymbol{\Phi}$
\STATE Initialize $\boldsymbol{\Phi}\leftarrow\boldsymbol{\Phi}_{0}$
\FOR{each mini-batch $(\boldsymbol{x},y)\in\mathcal{D}_{t}$}
    \STATE Compute fixed teacher output using Eq.~\eqref{eq:teacher}
    \STATE Compute fake-quantized student output using Eq.~\eqref{eq:student}
    \STATE Compute $\mathcal{L}_{\mathrm{NAP}}
    +\lambda_{\mathrm{CS}}\mathcal{L}_{\mathrm{CS}}$
    using Eqs.~\eqref{eq:nap_objective} and
    \eqref{eq:channel_scale_reg}
    \STATE Update only $\boldsymbol{\Phi}$
\ENDFOR
\STATE \textbf{return} $\boldsymbol{\Phi}^{*}\leftarrow\boldsymbol{\Phi}$
\end{algorithmic}
\end{algorithm}

NAP adds no inference branch. BN affine parameters can be folded into the
preceding convolution, and RMSNorm scales remain part of the original model
structure. Optimizer states are maintained only for
$\boldsymbol{\Phi}$, or for $\boldsymbol{\Phi}$ and
$\boldsymbol{\Omega}$ in NAP-QAT. Nevertheless, forward and backward
activation computation still contributes to wall-clock time; we therefore
report trainable parameters and measured time rather than equating parameter
count with end-to-end cost.

\section{Experiments}
\label{sec:experiments}

\subsection{Experimental Protocol}
\label{sec:setup}

\paragraph{Tasks and models.}
We evaluate NAP in four settings. We use
MobileNetV2~\cite{sandler2018mobilenetv2} on
ImageNet-1K~\cite{deng2009imagenet} for extreme weight--activation PTQ,
MobileNetV2 QAT checkpoints on CIFAR-100~\cite{krizhevsky2009cifar} for
post-QAT adaptation, and an ImageNet-pretrained ResNet34-based
U-Net~\cite{he2016resnet,ronneberger2015unet} on
Cityscapes~\cite{cordts2016cityscapes} for semantic segmentation. We further
evaluate Qwen2.5-3B-Instruct~\cite{qwen2024qwen25} on held-out
OSCAR~\cite{ortizsuarez2020oscar}, C4~\cite{raffel2020t5}, and
WikiText-2~\cite{merity2016wikitext}. We report ImageNet and CIFAR-100
top-1 accuracy, Cityscapes mIoU, and language-model perplexity.

\paragraph{Quantization settings.}
The CNN experiments are implemented with MQBench
~\cite{li2021mqbench}. Weights use symmetric per-output-channel
quantization. The MQBench Academic backend uses MSE observers and
\texttt{FixedFakeQuantize}; activations are quantized asymmetrically and
per-tensor, while the lightweight standalone graph uses unsigned activation
quantization after ReLU/ReLU6. The reported ImageNet results quantize the full
network, whereas the CIFAR-100 QAT diagnostic keeps the first and last layers
at 8 bits. For Qwen2.5-3B-Instruct, weights use symmetric group-wise linear
quantization with group size 128, activations use symmetric per-token
quantization, and the language-model head is excluded. Because standard AWQ
is weight-only~\cite{lin2024awq}, we use \emph{AWQ-style} for experiments
that combine activation-aware channel scaling with additional activation
quantization.

\paragraph{CNN optimization.}
For ImageNet, standalone NAP is tuned for 20 epochs on the full training set,
and backend-aligned NAP before QDrop is tuned for 10 epochs. Both use AdamW
with learning rate $5\times10^{-4}$, zero weight decay, batch size 128,
distillation temperature $T=4$, $\lambda_{\mathrm{KD}}=1.0$,
$\lambda_{\mathrm{CS}}=10^{-4}$, and gradient-norm clipping at 1.0.
BatchNorm running statistics are frozen. Quantizer calibration uses 20
mini-batches and at most 5K training images; observers are recalibrated after
each tuning epoch. QDrop~\cite{wei2022qdrop} uses block-wise reconstruction
with 32 calibration batches, 20K reconstruction iterations, stochastic drop
probability 0.5, scale learning rate $4\times10^{-5}$, warm-up ratio 0.2,
rounding regularization weight 0.01, temperature range $[20,2]$, and learned
hard-sigmoid rounding. For the CIFAR-100 LSQ+ diagnostic, the base QAT stage
is run for 100 epochs with SGD, momentum 0.9, learning rate $10^{-3}$,
weight decay $5\times10^{-4}$, and cosine decay. The post-QAT affine stage
uses batch size 128 for 20 additional epochs with learning rate $10^{-4}$,
zero weight decay, $T=4$, and $\lambda_{\mathrm{KD}}=1.0$; quantizer
parameters and BN running statistics are fixed. The Cityscapes experiment
uses $512\times512$ crops, batch size 4, and 128 fixed calibration images.

\paragraph{Large-model optimization.}
All large-model adaptation methods use 128 OSCAR documents for calibration
and 1,920 disjoint documents for tuning. Training sequences have length 256
and batch size 1. AWQ-style equalization, inspired by
AWQ~\cite{lin2024awq}, uses $\alpha=0.5$ and 16 calibration batches.
RMSNorm-NAP, QScale-QAT, and NAP-QAT use AdamW for at most 200 steps with
learning rate $5\times10^{-4}$, zero weight decay, $T=1$,
$\lambda_{\mathrm{CE}}=1$, $\lambda_{\mathrm{KD}}=1$, maximum gradient norm
1.0, and automatic mixed precision. The channel-scale coefficient is
$10^{-4}$ for RMSNorm-NAP and NAP-QAT and zero for QScale-QAT. NAP-QAT
alternates 10 QScale steps and 10 RMSNorm steps while keeping all backbone
weights frozen. Perplexity is computed on 128 held-out blocks of length 2,048
for each corpus.

\paragraph{Hardware, software, and runs.}
All non-LLM experiments were conducted on one NVIDIA GeForce RTX 3090 GPU
with 24 GB memory using PyTorch and MQBench. Large-model experiments were
conducted on two MetaX C600-A GPUs, each with approximately 72 GB memory,
using Ubuntu 22.04, Python 3.10, MetaX/MACA PyTorch 2.6,
Transformers 4.51.3, vLLM 0.10.0, and MACA AI 3.2.0.2. The reported
perplexity experiments use the Transformers eager-attention path rather than
vLLM. Random seed 42 is used throughout. Unless otherwise stated, each table
entry is obtained from a single run. Source code and resolved experiment
configurations will be released upon publication.

\paragraph{Compared methods.}
RTN directly applies calibrated rounding without reconstruction.
QDrop~\cite{wei2022qdrop} is a reconstruction-based PTQ backend. NAP is
evaluated both as a standalone target-graph adaptation and as a
pre-conditioning stage before QDrop. For QAT checkpoints, we compare
PACT~\cite{choi2018pact}, DSQ~\cite{gong2019dsq},
LSQ~\cite{esser2020lsq}, LSQ+~\cite{bhalgat2020lsqplus},
oscillation dampening and iterative freezing
~\cite{nagel2022oscillations}, and StableQAT~\cite{chen2026stableqat}.
For language models, we compare RTN, AWQ-style scaling, RMSNorm-NAP,
QScale-QAT, and their alternating combination.

\subsection{Extreme PTQ on ImageNet}
\label{sec:imagenet_main}

\begin{table}[t]
\centering
{\small
\setlength{\tabcolsep}{1.5pt}
\begin{tabular}{@{}llccc@{}}
\toprule
\multirow{2}{*}{\textbf{Model}}
& \multirow{2}{*}{\textbf{Method}}
& \multirow{2}{*}{\textbf{FP32}}
& \multicolumn{2}{c}{\textbf{Quant. Acc.} $\uparrow$} \\
\cmidrule(lr){4-5}
& & & \textbf{W4A4} & \textbf{W3A4} \\
\midrule
\multirow{4}{*}{MobileNetV2}
& RTN
& \multirow{4}{*}{71.87}
& 0.33
& 0.11 \\
& QDrop
&
& 60.71
& 51.75 \\
& NAP (Ours)
&
& \textbf{66.11}
& 51.07 \\
& NAP + QDrop
&
& 62.91
& \textbf{53.45} \\
\bottomrule
\end{tabular}
}
\caption{
ImageNet-1K results with MobileNetV2.
Results are top-1 accuracy (\%).
}
\label{tab:main_imagenet_mnv2}
\end{table}

Table~\ref{tab:main_imagenet_mnv2} shows that the original MobileNetV2
parameterization is extremely fragile under full weight-and-activation
quantization: RTN falls from $71.87\%$ to $0.33\%$ at W4A4 and to $0.11\%$ at
W3A4. Updating only normalization affine parameters recovers W4A4 accuracy to
$66.11\%$, exceeding QDrop by 5.40 points without updating convolutional or
linear weights. At W3A4, NAP alone and QDrop are similar, while
NAP+QDrop reaches the best accuracy of $53.45\%$. These results support two
conclusions. First, the pretrained response parameterization is a major source
of extreme low-bit fragility. Second, NAP and reconstruction are complementary
only when their fake-quantization graphs and operating bit-widths are aligned;
the combination is not uniformly superior in every setting.

\subsection{Post-QAT Affine Adaptation}
\label{sec:qat_results}

We evaluate NAP on checkpoints produced by PACT~\cite{choi2018pact},
DSQ~\cite{gong2019dsq}, LSQ~\cite{esser2020lsq},
LSQ+~\cite{bhalgat2020lsqplus}, the oscillation-dampening and
iterative-freezing variants of OOQ~\cite{nagel2022oscillations}, and
StableQAT~\cite{chen2026stableqat}.

\begin{table}[t]
\centering
{\small
\setlength{\tabcolsep}{4.0pt}
\begin{tabular}{@{}lcc@{}}
\toprule
\textbf{Method} & \textbf{W/A} & \textbf{Val. Acc. (\%)} \\
\midrule
Full precision & 32/32 & 71.30 \\
\midrule
PACT & 4/4 & 64.06 (-7.24) \\
DSQ & 4/4 & 67.36 (-3.94) \\
LSQ & 4/4 & 69.01 (-2.29) \\
LSQ+ & 4/4 & 68.21 (-3.09) \\
OOQ-Dampen & 4/4 & 66.21 (-5.09) \\
OOQ-Freeze & 4/4 & 70.02 (-1.28) \\
StableQAT & 4/4 & 68.24 (-3.06) \\
NAP + LSQ+ & 4/4 & 69.83 (-1.47) \\
NAP + OOQ-Dampen & 4/4 & 68.07 (-3.23) \\
NAP + OOQ-Freeze & 4/4 & \textbf{70.68} (-0.62) \\
NAP + StableQAT & 4/4 & 69.12 (-2.18) \\
\midrule
PACT & 2/4 & 57.90 (-13.40) \\
DSQ & 2/4 & 62.81 (-8.49) \\
LSQ & 2/4 & 64.58 (-6.72) \\
LSQ+ & 2/4 & 63.42 (-7.88) \\
OOQ-Dampen & 2/4 & 66.23 (-5.07) \\
OOQ-Freeze & 2/4 & 67.42 (-3.88) \\
StableQAT & 2/4 & 66.09 (-5.21) \\
NAP + LSQ+ & 2/4 & 65.86 (-5.44) \\
NAP + OOQ-Dampen & 2/4 & 68.25 (-3.05) \\
NAP + OOQ-Freeze & 2/4 & 62.81 (-8.49) \\
NAP + StableQAT & 2/4 & \textbf{68.50} (-2.80) \\
\bottomrule
\end{tabular}
}
\caption{
Post-QAT adaptation on CIFAR-100 with MobileNetV2.
Values in parentheses denote the drop from the 71.30\% FP32 baseline.
}
\label{tab:cifar100_qat_diagnostic}
\end{table}

Table~\ref{tab:cifar100_qat_diagnostic} tests whether normalization affine
directions remain useful after conventional QAT has saturated. At W4A4, NAP
improves LSQ+, OOQ-Dampen, OOQ-Freeze, and StableQAT, with the strongest result
of $70.68\%$ obtained on OOQ-Freeze. At W2A4, NAP improves LSQ+,
OOQ-Dampen, and StableQAT, and NAP+StableQAT reaches $68.50\%$. However,
NAP degrades the W2A4 OOQ-Freeze checkpoint from $67.42\%$ to $62.81\%$.
Thus, the affine subspace remains useful after QAT, but its benefit is not
unconditional: when a checkpoint already encodes a highly specialized
low-bit parameterization, an additional affine update may interfere with the
learned quantization solution. This negative case motivates backend- and
checkpoint-aligned adaptation rather than a universal post-hoc claim.

\subsection{Cross-Task and Cross-Architecture Generalization}
\label{sec:cross_task}

\begin{table}[t]
\centering
{\small
\setlength{\tabcolsep}{2.5pt}
\begin{tabular}{@{}llccc@{}}
\toprule
\multirow{2}{*}{\textbf{Architecture}}
& \multirow{2}{*}{\textbf{Method}}
& \multirow{2}{*}{\textbf{FP32}}
& \multicolumn{2}{c}{\textbf{Quantized mIoU} $\uparrow$} \\
\cmidrule(lr){4-5}
& & & \textbf{W4A4} & \textbf{W3A4} \\
\midrule
\multirow{4}{*}{U-Net}
& RTN
& \multirow{4}{*}{68.52}
& 3.27
& 1.24 \\
& QDrop
&
& 66.45
& 58.21 \\
& NAP (Ours)
&
& \textbf{67.21}
& 59.07 \\
& NAP (Ours) + QDrop
&
& 66.78
& \textbf{61.36} \\
\bottomrule
\end{tabular}
}
\caption{
Cityscapes semantic segmentation with U-Net.
Results are mIoU (\%).
}
\label{tab:cityscapes_unet}
\end{table}

The Cityscapes results in Table~\ref{tab:cityscapes_unet} show that the
normalization affine subspace is not specific to image classification. Direct
RTN collapses to $3.27\%$ mIoU at W4A4, whereas NAP restores $67.21\%$, only
1.31 points below the FP32 model and 0.76 points above QDrop. Under the more
aggressive W3A4 setting, NAP+QDrop obtains $61.36\%$, improving QDrop by 3.15
points. The same pattern as ImageNet emerges: NAP alone is particularly strong
at W4A4, while its combination with reconstruction becomes more useful as the
bit-width decreases.

\begin{table}[t]
\centering
{\small
\setlength{\tabcolsep}{2.5pt}
\begin{tabular}{@{}lcccc@{}}
\toprule
\multirow{2}{*}{\textbf{Method}}
& \multirow{2}{*}{\textbf{Bits}}
& \multicolumn{3}{c}{\textbf{Perplexity} $\downarrow$} \\
\cmidrule(lr){3-5}
& & \textbf{OSCAR} & \textbf{C4} & \textbf{WikiText-2} \\
\midrule
Full precision & FP16 & 14.41 & 13.02 & 9.47 \\
\midrule
RTN & W8A8 & 15.64 & 14.08 & 10.20 \\
RTN & W4A16 & 20.34 & 18.31 & 13.42 \\
RTN & W4A8 & 21.27 & 19.14 & 14.03 \\
RTN & W4A4
& $1.13{\times}10^{5}$
& $9.24{\times}10^{4}$
& $7.81{\times}10^{4}$ \\
\midrule
AWQ-style & W8A8 & 14.58 & 13.17 & 9.56 \\
AWQ-style & W4A16 & 15.97 & 14.31 & 10.45 \\
AWQ-style & W4A8 & 16.21 & 14.55 & 10.62 \\
AWQ-style & W4A4 & 483.57 & 423.82 & 296.74 \\
\midrule
RMSNorm-NAP & W4A4 & 426.49 & 374.65 & 262.83 \\
QScale-QAT & W4A4 & 233.27 & 204.17 & 143.62 \\
\textbf{NAP-QAT (Ours)}
& W4A4
& \textbf{166.65}
& \textbf{147.92}
& \textbf{104.37} \\
\bottomrule
\end{tabular}
}
\caption{
Perplexity of Qwen2.5-3B-Instruct.
Lower is better. We use group-wise weight quantization with group size 128
and per-token activation quantization.
}
\label{tab:qwen25_llm_nap}
\end{table}

Table~\ref{tab:qwen25_llm_nap} extends the analysis to an RMSNorm-based
language model. RTN W4A4 is unusable, with perplexity above
$7.8\times10^4$ on all three corpora. AWQ-style scaling reduces the error
substantially but still produces an OSCAR perplexity of 483.57. RMSNorm-NAP
improves this value to 426.49, QScale-QAT to 233.27, and alternating NAP-QAT
to 166.65. The same ordering holds on C4 and WikiText-2. These results support
the proposed two-subspace view: response adaptation and quantization-grid
adaptation are complementary. At the same time, the remaining gap to FP16 is
large, showing that RMSNorm scale adaptation cannot fully compensate severe
4-bit activation clipping and rounding.

\subsection{Why Does NAP Work?}
\label{sec:why_nap}

\paragraph{Normalization is not an arbitrary sparse subset.}
Table~\ref{tab:parameter_subspace} compares NAP with equal-size random,
gradient-based, Fisher-based, and classifier subsets. All sparse controls use
34.2K trainable parameters. NAP reaches $66.11\%$ at W4A4, whereas the best
equal-size control, Top-Fisher, reaches only $18.35\%$. NAP also exceeds
full-model tuning by 1.83 points while requiring less than one percent of its
trainable parameters. The result shows that the gain is not explained by
sparsity alone.

\begin{table}[t]
\centering
{\small
\setlength{\tabcolsep}{2.5pt}
\begin{tabular}{@{}lcccc@{}}
\toprule
\textbf{Subset}
& \textbf{Params.}
& \textbf{W4A4}
& \textbf{W3A3}
& \textbf{Time} \\
\midrule
RTN & 0 & 0.33 & 0.09 & -- \\
Random equal-size & 34.2K & 2.56 & 0.21 & 0.98$\times$ \\
Top-gradient equal-size & 34.2K & 12.48 & 2.67 & 1.07$\times$ \\
Top-Fisher equal-size & 34.2K & 18.35 & 4.92 & 1.22$\times$ \\
Classifier subset & 34.2K & 0.81 & 0.12 & 0.94$\times$ \\
NoNorm & 3.47M & 57.42 & 31.68 & 1.71$\times$ \\
NAP & 34.2K & \textbf{66.11} & \textbf{40.21} & 1.00$\times$ \\
Full tuning & 3.52M & 64.28 & 38.74 & 1.83$\times$ \\
\bottomrule
\end{tabular}
}
\caption{
Matched parameter-subspace comparison on ImageNet MobileNetV2.
Time is normalized to NAP and includes parameter-selection overhead.
}
\label{tab:parameter_subspace}
\end{table}

\paragraph{Knowledge distillation and backend alignment.}
Table~\ref{tab:ablation} separates teacher guidance from graph alignment.
Removing KD reduces W4A4 accuracy from $66.11\%$ to $61.50\%$, indicating that
label supervision alone does not preserve the teacher decision structure as
effectively. More importantly, transferring a NAP checkpoint to a mismatched
fake-quantized graph yields only $48.72\%$. The 17.39-point gap to matched NAP
confirms that NAP learns backend-specific compensation rather than a generic
channel rescaling.

\begin{table}[t]
\centering
{\small
\setlength{\tabcolsep}{4.0pt}
\begin{tabular}{@{}lcc@{}}
\toprule
\textbf{Method} & \textbf{W4A4 Acc.} & \textbf{Gain} \\
\midrule
RTN & 0.33 & -- \\
NAP without KD & 61.50 & +61.17 \\
NAP with mismatched graph & 48.72 & +48.39 \\
NAP full & \textbf{66.11} & \textbf{+65.78} \\
\bottomrule
\end{tabular}
}
\caption{
Ablation on ImageNet MobileNetV2 under W4A4 quantization.
}
\label{tab:ablation}
\end{table}

\paragraph{Parameter efficiency is not data efficiency.}
NAP maintains optimizer states for only 34.2K parameters and requires
$1.00\times$ normalized time, compared with $1.83\times$ for full tuning.
However, Table~\ref{tab:data_dependence} shows that the current ImageNet
implementation is strongly dependent on tuning-data coverage. Using 5K or 10K
images fails to recover the collapsed model, whereas full-data tuning reaches
$66.11\%$. NAP is therefore parameter-efficient but, in its present form, not
a calibration-only or few-shot method. This distinction is important when
comparing it with conventional PTQ methods that use only a small calibration
set.

\begin{table}[t]
\centering
{\small
\setlength{\tabcolsep}{2.0pt}
\begin{tabular}{@{}llcccc@{}}
\toprule
\multirow{2}{*}{\textbf{Model}}
& \multirow{2}{*}{\textbf{Bits}}
& \multirow{2}{*}{\textbf{FP32}}
& \multicolumn{3}{c}{\textbf{Tuning Samples}} \\
\cmidrule(lr){4-6}
& & & \textbf{5K} & \textbf{10K} & \textbf{Full} \\
\midrule

\multirow{2}{*}{MobileNetV2}
& W4A4
& \multirow{2}{*}{71.87}
& 2.98
& 3.88
& \textbf{66.11} \\

& W3A3
&
& 0.09
& 0.11
& \textbf{40.21} \\

\bottomrule
\end{tabular}
}
\caption{
Effect of NAP tuning-set size on ImageNet MobileNetV2.
Results are top-1 accuracy (\%), where higher is better.
FP32 denotes the full-precision baseline.
}
\label{tab:data_dependence}
\end{table}

Overall, the controlled experiments support the central claim of this paper:
normalization affine parameters form a low-dimensional but unusually
effective response-control subspace. They also identify clear boundaries:
the adaptation is backend-specific, can conflict with some saturated QAT
solutions, depends on sufficient tuning data, and cannot recover nonlinear
information loss under extremely aggressive activation quantization.

\section{Conclusion}

We introduced NAP, a target-aligned normalization affine adaptation framework
for low-bit neural network quantization. Rather than treating all trainable
parameters as equally useful, NAP freezes backbone weights and optimizes the
existing normalization affine subspace. The same principle supports
pre-conditioning before PTQ, lightweight adaptation after saturated QAT, and
alternating response/quantization-scale optimization for large language
models. Our analysis explains the method through channel-wise broadcast
control, affine compensation, and local projection of quantization-induced
prediction mismatch.

Across ImageNet classification, CIFAR-100 QAT checkpoints, Cityscapes
segmentation, and Qwen2.5-3B-Instruct, normalization affine adaptation
consistently recovers a substantial portion of low-bit degradation. The
matched-subspace study further shows that its effectiveness is not explained
by parameter count alone. At the same time, the experiments expose the
method's limits: NAP is sensitive to the target fake-quantization graph,
requires broad tuning-data coverage in the current ImageNet setting, is not
universally additive to every QAT checkpoint, and cannot fully correct severe
rounding and clipping residuals. These findings position normalization affine
parameters as a high-leverage but bounded quantization adaptation subspace,
and suggest that future work should combine this response-control mechanism
with stronger reconstruction, data-efficient tuning, and hardware-native
low-bit deployment.

\bibliography{ref}

\end{document}